\pdfoutput=1

\documentclass[11pt]{article}
\usepackage{microtype}
\usepackage{acl}
\usepackage{soul}
\usepackage{bm}
\usepackage{amsmath}
\usepackage{amssymb}
\usepackage{times}
\usepackage{latexsym}
\usepackage{graphicx}
\usepackage{subfigure}

\usepackage{caption}
\usepackage{mathrsfs}
\usepackage[T1]{fontenc}

\usepackage[utf8]{inputenc}

\usepackage{microtype}

\usepackage{graphicx}

\usepackage{multirow}
\usepackage{enumitem}

\usepackage[T1]{tipa}
\usepackage{makecell}

\usepackage{url}
\usepackage{tablefootnote}

\newcommand{\vtheta}{{\boldsymbol \theta}}
\newcommand{\vocab}{\mathcal{V}}

\usepackage{color}
\usepackage{tikz}
\usepackage[edges]{forest}
\definecolor{hiddendraw}{RGB}{205, 44, 36}
\definecolor{hidden-blue}{RGB}{194,232,247}
\definecolor{hidden-orange}{RGB}{243,202,120}
\definecolor{hidden-yellow}{RGB}{242,244,193}
\tikzstyle{mybox}=[
    rectangle,
    draw=hiddendraw,
    rounded corners,
    text opacity=1,
    minimum height=1.5em,
    minimum width=5em,
    inner sep=2pt,
    align=center,
    fill opacity=.5,
    ]

\usepackage{amssymb}
\usepackage{pifont}
\usepackage{amsthm,amsmath,amssymb}
\usepackage{mathrsfs}

\usepackage{amsthm}

\usepackage{multirow} 
\usepackage{booktabs} 

\title{Cognitive Mirage: A Review of Hallucinations in Large Language Models}

\author{Hongbin Ye, Tong Liu, Aijia Zhang, Wei Hua, Weiqiang Jia\\
  Zhejiang Lab\\
  \texttt{\{yehongbin,jiaweiqiang\}@zhejianglab.com}
  }

\begin{document}

\maketitle

\begin{abstract}

As large language models continue to develop in the field of AI, text generation systems are susceptible to a worrisome phenomenon known as \textit{hallucination}. In this study, we summarize recent compelling insights into hallucinations in LLMs. We present a novel taxonomy of hallucinations from various text generation tasks, thus provide theoretical insights, detection methods and improvement approaches. Based on this, future research directions are proposed. Our contribution are threefold: (1) We provide a detailed and complete taxonomy for hallucinations appearing in text generation tasks; (2) We provide theoretical analyses of hallucinations in LLMs and provide existing detection and improvement methods; (3) We propose several research directions that can be developed in the future. 
As hallucinations garner significant attention from the community, we will maintain updates on relevant research progress\footnote{\url{https://github.com/hongbinye/Cognitive-Mirage-Hallucinations-in-LLMs}}.

\end{abstract}

\section{Introduction}

\begin{figure}
    \centering
    \resizebox{.49\textwidth}{!}{
    \includegraphics{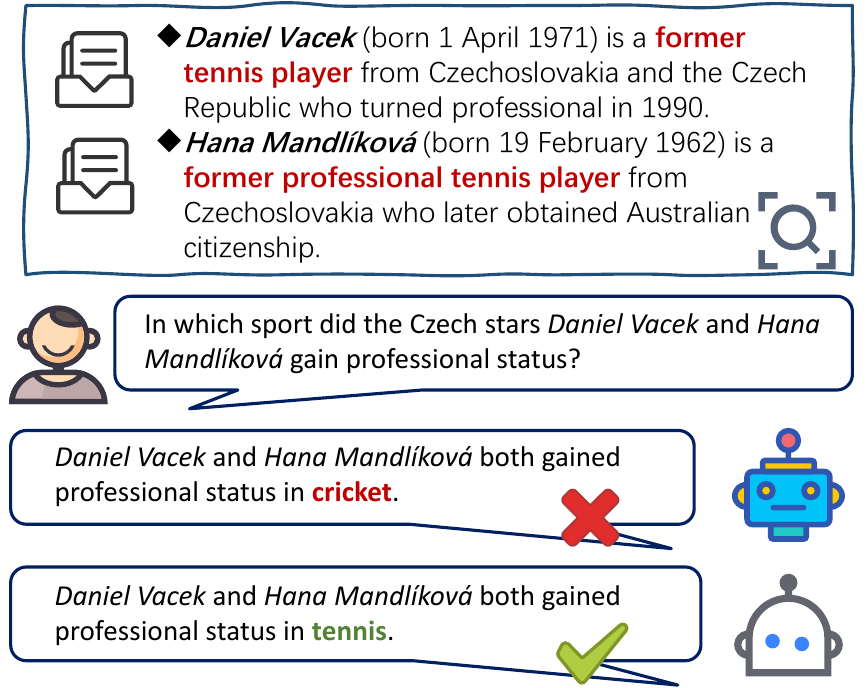}}
    \caption{
    Illustration of Hallucination in LLMs. While the initial response appears fluent, it fails to align with the world knowledge retrieved from the external knowledge base.}
    \label{fig:intro}
\end{figure}

In the ever-evolving realm of large language models (LLMs), a constellation of innovative creations has emerged, such as GPT-3 \citep{DBLP:conf/nips/BrownMRSKDNSSAA20}, InstructGPT \citep{DBLP:conf/nips/Ouyang0JAWMZASR22}, FLAN \citep{DBLP:conf/iclr/WeiBZGYLDDL22}, PaLM \citep{DBLP:journals/corr/abs-2204-02311}, LLaMA \citep{DBLP:journals/corr/abs-2302-13971} and other notable contributors \citep{DBLP:journals/corr/abs-2204-05862,DBLP:journals/corr/abs-2205-01068,DBLP:conf/iclr/ZengLDWL0YXZXTM23,DBLP:journals/corr/abs-2304-12244}.
These models implicitly encode global knowledge within their parameters during the pre-training phase
\citep{DBLP:journals/aiopen/HanZDGLHQYZZHHJ21,DBLP:conf/acl/0009C23}, offering valuable insights as knowledge repositories for downstream  tasks \citep{DBLP:conf/acl/PuD23,DBLP:conf/nips/KojimaGRMI22,DBLP:conf/nips/Wei0SBIXCLZ22}.
Nevertheless, the generalization of knowledge can result in \textit{memory distortion}, an inherent limitation that may give rise to potential inaccuracies \citep{DBLP:journals/corr/abs-2305-14002}.
Moreover, their ability to represent knowledge is constrained by model scale and faces challenges in addressing long-tailed knowledge problems \citep{DBLP:conf/icml/KandpalDRWR23,DBLP:conf/acl/MallenAZDKH23}.
While the privacy and timeliness of data in the real world \citep{DBLP:journals/corr/abs-2203-05115,DBLP:journals/corr/abs-2301-12652} unfortunately exacerbate this problem, leaving models difficult to maintain a comprehensive and up-to-date understanding of the facts.
These challenges present a serious obstacle to the reliability of LLMs, which we refer to as \textit{hallucination}. \citep{DBLP:journals/csur/YuZLHWJJ22}.
A prominent example of this drawback is that models typically generate statements that appear reasonable but are either cognitively irrelevant or factually incorrect.
In light of this observation, hallucinations remain a critical challenge in medical \citep{DBLP:journals/corr/abs-2304-13714,DBLP:journals/corr/abs-2307-15343}, financial \citep{DBLP:journals/corr/abs-2306-03823} and other knowledge-intensive fields due to the exacting accuracy requirements.
Particularly, the applications for legal case drafting showcase plausible interpretation as an aggregation of diverse subjective perspectives \citep{DBLP:journals/corr/abs-2306-11520}.

\begin{figure*}
    \centering
    \resizebox{.99\textwidth}{!}{
    \includegraphics{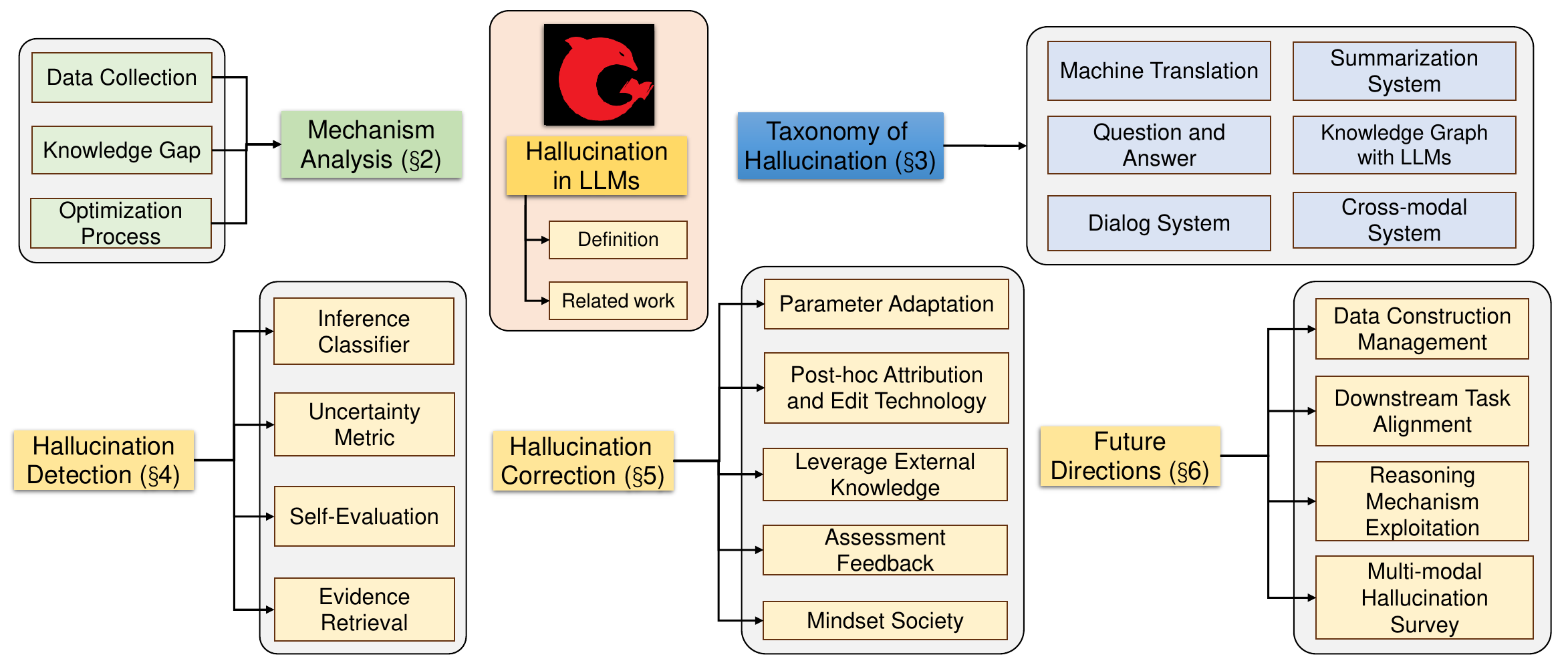}}
    \caption{
    The overview structure of this review. We firstly analyze three crucial factors that contribute to hallucinations and refine the categorization of hallucinations across text generation tasks. Subsequently, we dutifully report current methods for detecting and mitigating hallucinations. Finally, we propose several potential research directions to address evolving problems of hallucinations.}
    \label{fig:overview}
\end{figure*}

\paragraph{Definition of Hallucination.} 
As depicted in Figure~\ref{fig:intro},  \textit{hallucination} refers to the generation of texts or responses that exhibit grammatical correctness, fluency, and authenticity, but deviate from the provided source inputs (\textit{faithfulness}) or do not align with factual accuracy (\textit{factualness}) \citep{DBLP:journals/csur/JiLFYSXIBMF23}.
In traditional NLP tasks \citep{DBLP:conf/acl/MaynezNBM20}, hallucinations are often synonymous with \textit{faithfulness}: conflicting information leads to \textit{Intrinsic Hallucination}, i.e., LMs conflict with the input information when generating a response; Conversely, generating ambiguous supplementary information may leads to \textit{Extrinsic Hallucination}, i.e., LMs produce personal names, historical events, or technical documents that is challenging to verify.
LLMs-oriented hallucinations instead prioritize \textit{factualness}, focusing on whether the result can be evidenced or negated by reference to external facts in the real world.
Uncritical trust in LLMs can give rise to a phenomenon \textbf{Cognitive Mirage}, contributing to misguided decision-making and a cascade of unintended consequences \citep{DBLP:journals/corr/abs-2305-13534}.

\paragraph{Present work} 
To effectively control the risk of hallucinations, we summarize recent progress in hallucination theories and solutions in this paper.
We propose to organize relevant work by a comprehensive survey (Figure \ref{fig:overview}):

\begin{itemize}
    \item \textbf{Theoretical insight and mechanism analysis}. 
    We provide in-depth theoretical and mechanism analysis from three typical perspectives: data collection, knowledge gap and optimization process, reviewing the recent and relevant theories related to hallucinations (\S \ref{analysis}). 
    \item \textbf{Taxonomy of hallucination in LLMs}. 
    We conduct the comprehensive review of hallucination in LLMs together with task axis. 
    We review the task-specific benchmarks with a comprehensive comparison and summary (\S \ref{taxonomy}).
    \item \textbf{Wide coverage on emerging hallucination detection and correction methods}. 
    We propose a comprehensive investigation into the proactive detection (\S \ref{detection}) and mitigation of hallucinations (\S \ref{correction}) in the era of LLMs. This is critical to study the most popular techniques for inspiring future research directions (\S \ref{future work}).

\end{itemize}

\paragraph{Related work}
As this topic is relatively nascent, only a few surveys exist.
Closest to our work, \citet{DBLP:journals/csur/JiLFYSXIBMF23}  analyzes  hallucinatory content in task-specific research progress, which focuses on early works in natural language generation field.
Currently there are significant efforts to address hallucination in LLMs.
\citet{DBLP:journals/corr/abs-2307-12966} covers methods for effectively collecting high-quality instructions for LLM alignment, including the use of NLP benchmarks, human annotations, and leveraging strong LLMs.
\citet{DBLP:journals/corr/abs-2308-03188} discusses self-correcting methods where LLM itself is prompted or guided to correct the hallucinations from its own outputs.
Despite some benchmarks \citep{DBLP:conf/acl/LinHE22,DBLP:journals/corr/abs-2305-11747,DBLP:journals/corr/abs-2308-02357} is constructed to evaluate whether LLMs are able to generate factual responses, these works scattered among various tasks have not been systematically reviewed and analyzed.
Different from those surveys, in this paper, we conduct a literature review on hallucinations in LLMs, hoping to systematically understand the methodologies, compare different methods and inspire new ideas.

\section{Mechanism Analysis}
\label{analysis}
For the sake of clean exposition, this section provides theoretical insight into mechanism analysis for hallucinations in LLMs.
As a regular LLM, the generative objective is modeled by a parameterized probabilistic model $p_{gen}$, and sampled to predict the next token in the sentence, thus generating the entire sentence:
\begin{align}
p_{gen}(y_i)=\mathcal{F}_\vtheta(\mathcal{I},\mathcal{D},x,y_{i<})
\end{align}
where $y_i$ represents probable tokens at each step that can be selected by beam search from vocabulary $\vocab$. Note that the instructions $\mathcal{I}$ utilize a variety of predefined templates according to different tasks \citep{DBLP:conf/acl/YinVLJXW23}. Multifarious and high-quality in-context demonstrations $\mathcal{D}$ are aimed at providing analogy samples to reduce the cost of adapting models to new tasks \citep{DBLP:conf/naacl/ChenDPMISK22}.
Parameters $\vtheta$ implicitly memorize corpus knowledge through diverse architectural $\mathcal{F}$ such as decoder-only, encoder-only, or encoder-decoder LLMs.
As LLM-based systems can exhibit a variety of hallucinations, we summarise three primary mechanism types for theoretical analysis, and each mechanism is correlated with a distinct training factor.

\paragraph{Data Collection}
The parameters are implicitly stored within model as a priori knowledge acquired  during pre-training process. 
Given the varying quality and range of knowledge within  pre-trained corpus, the  information incorporated into the LLMs may be incomplete or outdated.
In cases where pertinent memories are unavailable, the LLM's performance may deteriorates, resorting to rudimentary corpus-based heuristics that rely on term frequencies to render judgements \citep{DBLP:journals/corr/abs-2305-14552}.
Another bias stems from the capacity for contextual learning \citep{DBLP:conf/nips/ChanSLWSRMH22}  when a few demonstrations are introduced as input to the prefix context. 
Previous research \citep{DBLP:conf/acl/WangWZL023,DBLP:conf/acl/LuBM0S22} has demonstrated that the acquisition of knowledge through model learning demonstrations depends on disparities in label categories and the order of demonstration samples.
Likewise, multilingual LLMs encounter challenges related to hallucinations, particularly in handling language pairs with limited resources or non-English translations \citep{DBLP:journals/corr/abs-2303-16104}.
Furthermore,  cutting-edge Large Vision-Language Models (LVLMs) exhibit instances of hallucinating common objects within visual instructional datasets and prone to objects that frequently co-occur in the same image \citep{DBLP:journals/corr/abs-2304-08485,DBLP:journals/corr/abs-2305-10355}.

\paragraph{Knowledge Gap}

Knowledge gaps are typically attributed to differences in input format between the pre-training and fine-tuning stages \citep{DBLP:journals/corr/abs-2304-10513}.
Even when considering the automatic updating of textual knowledge bases, the output can deviate from the expected corrections  \citep{DBLP:conf/acl/HuangCJ23}.
For example, questions often do not align effectively with stored knowledge, and the available information remains unknown until the questions are presented.
This knowledge gap poses thorny challenges in balancing memory with retrieved evidence, which is construed as a passive defense mechanism against the misuse of retrieval
\citep{DBLP:conf/acl/GaoDPCCFZLLJG23}.
To in-depth analyses this issue,
\citet{DBLP:journals/corr/abs-2305-13669} and \citet{DBLP:journals/corr/abs-2307-09476} propose that disregarding retrieved evidence introduces biased model knowledge, while mis-covering and over-thinking disrupt model behavior.
Furthermore, in scenarios where a cache component is utilized to offer historical memory during training \citep{DBLP:journals/corr/abs-2305-04782}, the model also experiences inconsistency between the present hidden state and the hidden state stored in the cache.

\paragraph{Optimization Process}

The maximum likelihood estimation and teacher-forcing training have the potential to result in a phenomenon known as \textit{stochastic parroting} \citep{DBLP:conf/acl/ChiesurinDCEPRK23}, wherein the model is prompted to imitate the training data without comprehension \citep{DBLP:conf/acl/KangH20}.
Specifically, exposure bias between the training and testing stages have been demonstrated to lead to hallucinations within LLMs, particularly when generating lengthy responses \citep{DBLP:conf/acl/WangS20}.
Besides, sampling techniques characterized by high uncertainty \citep{DBLP:conf/nips/LeePXPFSC22}, such as top-p and top-k, exacerbate the issue of hallucination.
Furthermore, \citet{DBLP:journals/corr/abs-2305-13534} observes that LLMs tend to produce snowballing hallucinations to maintain coherence with earlier hallucinations, and even when directed with prompts as "Let's think step by step", they still generate ineffectual chains of reasoning \citep{DBLP:conf/nips/KojimaGRMI22}.

\begin{table*}[!t]
\small
\centering
\resizebox{0.97\textwidth}{!}{
\begin{tabular}{p{0.12\linewidth}|p{0.1\linewidth}p{0.1\linewidth}p{0.15\linewidth}p{0.2\linewidth}p{0.17\linewidth}}
\toprule
 \textbf{Paper} & \textbf{Task} &  \textbf{Architecture} & \textbf{Resources} & \textbf{Hallucination Types} & \textbf{Research Method}\\
\midrule
 \citet{DBLP:conf/naacl/RaunakMJ21} & Machine Translation & Enc-Dec  & IWSLT-2014  & Under perturbation, Natural hallucination & Source perturbation   \\
\midrule
 \citet{DBLP:conf/eacl/GuerreiroVM23} & Machine Translation & Enc-Dec  & WMT2018   & Oscillatory hallucination, Largely fluent hallucination & Consider a natural scenario \\
 \midrule
 \citet{DBLP:journals/corr/abs-2305-11746} & Machine Translation & Enc-Dec  & FLORES-200, Jigsaw, Wikipedia  & Full hallucination, Partial hallucination, Word-level hallucination & Introduce pathology detection   \\
  \midrule
 \citet{DBLP:journals/corr/abs-2305-14224} & Multilingual Seq2seq & Enc-Dec  & XQuAD, TyDi, XNLI, XL-Sum, MASSIVE  & Source language hallucination &  Evaluate source language hallucination   \\
\midrule
 \citet{DBLP:conf/acl/LinHE22} & Question and Answer & Enc-Dec, Only-Dec  & TruthfulQA  &  Imitative falsehoods & Cause imitative falsehoods   \\
  \midrule
 \citet{DBLP:journals/corr/abs-2304-10513} &  Question and Answer & Only-Dec  & HotpotQA, BoolQ  & Comprehension, Factualness, Specificity, Inference Hallucination  & Manual analysis of responses   \\
  \midrule
 \citet{DBLP:journals/corr/abs-2307-16877} & Question and Answer & Enc-Dec, Only-Dec  &  NQ, HotpotQA,  TopiOCQA  & Semantic equivalence, Symbolic equivalence, Intrinsic ambiguity, Granularity discrepancies, Incomplete, Enumeration, Satisfactory Subset  & Evaluate retrieval augmented QA   \\
 \midrule
 \citet{DBLP:journals/corr/abs-2307-15343} & Question and Answer & Only-Dec  & MEDMCQA, Headqa, USMILE, Medqa, Pubmed  & Reasoning hallucination, Memory-based hallucination  & Medical benchmark \textit{Med-HALT}  \\
 \midrule
\citet{DBLP:conf/naacl/DziriMYZR22} & Dialog $\quad$ System & Enc-Dec, Only-Dec  & WoW, CMU-DOG, TopicalChat   & Hallucination, Partial hallucination, Generic, Uncooperative & Infer exclusively from the knowledge-snippet   \\
\midrule
\citet{DBLP:conf/emnlp/DasSS22} &  Dialog $\quad$ System & Only-Dec  & OpenDialKG  & Extrinsic-Soft/Hard/ Grouped, Intrinsic-Soft/ Hard/Repetitive, History Corrupted & Analyze entity-level fact hallucination  \\
\midrule
\citet{DBLP:journals/tacl/DziriKMZYPR22} & Dialog $\quad$ System & Enc-Dec, Only-Dec  & WoW  & Hallucination, Generic, Uncooperativeness & Hallucination-free  benchmark \textit{FaithDial}  \\
\midrule
 \citet{DBLP:journals/tacl/DziriRLR22} & Dialog $\quad$ System & Enc-Dec, Only-Enc,  Only-Dec  & WoW, CMU-DOG, TopicalChat  & Fully attributable, Not attributable, Generic & Knowledge-grounded interaction benchmark \textit{Begin} \\
\midrule
\citet{DBLP:conf/aaai/SunSGRRR23} & Dialog $\quad$ System &  Enc-Dec, Only-Dec  & WoW  & Intrinsic hallucination, Extrinsic hallucination &  Sample responses for conversation  \\
 \midrule
 \citet{DBLP:conf/acl/TamMZKBR23} & Summarization System & Enc-Dec, Only-Dec  & CNN/DM, XSum  & Factually inconsistent summaries & Generate summaries from given models   \\
 \midrule
 \citet{DBLP:conf/acl/CaoDC22} & Summarization System & Enc-Dec, Only-Dec  & MENT  & Non-hallucinated, Factual hallucination, Non-factual hallucination, Intrinsic hallucination & Label factual entities from summarizations  \\
 \midrule
 \citet{DBLP:conf/www/ShenLFRBN23} & Summarization System & Enc-Dec, Only-Enc  & NHNet  & News headline hallucination & Majority vote of journalism degree holders   \\
 \midrule
 \citet{DBLP:journals/corr/abs-2305-13632} & Summarization System & Multiple ADapters  & XL-Sum  & Intrinsic hallucination, Extrinsic hallucination &  In a cross-lingual transfer setting   \\
 \midrule
 \citet{DBLP:journals/corr/abs-2306-09296} & Knowledge-based text generation & Enc-Dec, Only-Dec  &  Encyclopedic, ETC   & Knowledge hallucination &  Evaluate knowledge creating ability given known facts  \\
 \midrule
\citet{DBLP:journals/corr/abs-2308-02357} & Knowledge graph generation &  Only-Dec  &  TekGen, WebNLG   & Subject hallucination,  relation hallucination, object hallucination &  Ontology driven KGC benchmark \textit{Text2KGBench}  \\
 \midrule
\citet{DBLP:journals/corr/abs-2305-10355} & Visual Question Answer & Enc-Dec  & MSCOCO  & Object hallucination & Caption hallucination assessment   \\
 \bottomrule
 \end{tabular}
}
 \caption{List of Representative Hallucination}
 \label{tab:representative-hallucination}
 \end{table*}

\section{Taxonomy of Hallucination}
\label{taxonomy}

\tikzstyle{leaf}=[mybox,minimum height=1.5em,
fill=hidden-orange!60, text width=20em,  text=black,align=left,font=\scriptsize,
inner xsep=2pt,
inner ysep=4pt,
]

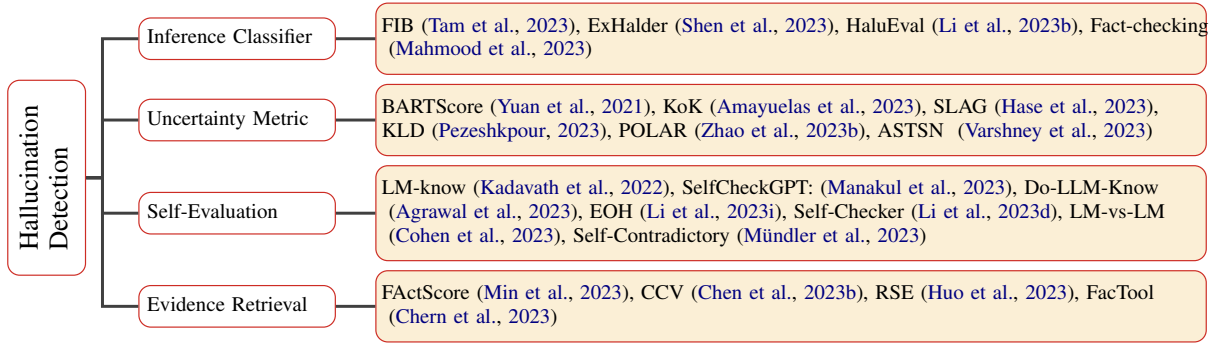
\begin{figure*}[!ht]
  \centering
  \resizebox{\textwidth}{!}{
\begin{forest}
  forked edges,
  for tree={
  grow=east,
  reversed=true,
  anchor=base west,
  parent anchor=east,
  child anchor=west,
  base=left,
  font=\small,
  rectangle,
  draw=hiddendraw,
  rounded corners,align=left,
  minimum width=6em,
    edge+={darkgray, line width=1pt},
s sep=3pt,
inner xsep=1pt,
inner ysep=3pt,
ver/.style={rotate=90, child anchor=north, parent anchor=south, anchor=center},
  },
  where level=1{text width=5.5em,font=\scriptsize,}{},
  where level=2{text width=4em,font=\scriptsize,}{},
  [Hallucination \\ Detection   , ver 
        [Inference Classifier  
            [ FIB~\citep{DBLP:conf/acl/TamMZKBR23}{,}
            ExHalder~\citep{DBLP:conf/www/ShenLFRBN23}{,}
            HaluEval~\citep{DBLP:journals/corr/abs-2305-11747}{,}
            Fact-checking \\ ~\citep{DBLP:journals/corr/abs-2307-14634},
            ,leaf,text width=25em]
        ]
        [Uncertainty Metric
            [ BARTScore~\citep{DBLP:conf/nips/YuanNL21}{,}
            KoK~\citep{DBLP:journals/corr/abs-2305-13712}{,}
            SLAG~\citep{DBLP:conf/eacl/HaseDCLKSBI23}{,} \\
            KLD~\citep{DBLP:journals/corr/abs-2306-06264}{,} 
            POLAR~\citep{zhao2023llm}{,} 
            ASTSN ~\citep{DBLP:journals/corr/abs-2307-03987},
            ,leaf,text width=25em]
        ]
        [Self-Evaluation
            [ LM-know~\citep{DBLP:journals/corr/abs-2207-05221}{,}
            SelfCheckGPT:~\citep{DBLP:journals/corr/abs-2303-08896}{,}
            Do-LLM-Know \\ ~\citep{DBLP:journals/corr/abs-2305-18248}{,}
            EOH~\citep{DBLP:journals/corr/abs-2305-10355}{,}
            Self-Checker~\citep{DBLP:journals/corr/abs-2305-14623}{,}
            LM-vs-LM \\~\citep{DBLP:journals/corr/abs-2305-13281}{,}
            Self-Contradictory~\citep{DBLP:journals/corr/abs-2305-15852},
            ,leaf,text width=25em]
        ]
        [Evidence Retrieval
            [ FActScore~\citep{DBLP:journals/corr/abs-2305-14251}{,}
            CCV~\citep{DBLP:journals/corr/abs-2305-11859}{,}
            RSE~\citep{DBLP:journals/corr/abs-2306-13781}{,}
            FacTool \\ ~\citep{DBLP:journals/corr/abs-2307-13528}, 
            ,leaf,text width=25em]
        ]
    ]
\end{forest}
}
\caption{Taxonomy of Hallucination Detection.}
\label{fig:detection}
\end{figure*}

In this paper, we mainly consider representative hallucinations, which are widely observed in various downstream tasks, i.e. \emph{Machine Translation}, \emph{Question and Answer}, \emph{Dialog System}, \emph{Summarization System}, \emph{Knowledge graph with LLMs}, and \emph{Visual Question Answer}. 
As shown in Table~\ref{tab:representative-hallucination}, these hallucinations are identified complex taxonomy in numerous mainstream tasks associated with LLMs.
In the following sections, we will introduce representative  types of hallucinations to be resolved.

\noindent $\bullet$  \textbf{Machine Translation.}
Since perturbations (e.g., spellings or capital errors) can induce hallucinations reliably, traditional machine translation models tend to validate instances memorised by the model when subjected to perturbations
\citep{DBLP:journals/corr/abs-2303-01911,DBLP:journals/corr/abs-2302-09210}.
It is worth noting that hallucinations generated by LLMs are mainly translation off-target, over-generation, or failed translation attempts \citep{DBLP:journals/corr/abs-2303-16104}.
While in low-resource language setting, most models exhibit subpar performance due to the lack of annotated data \citep{DBLP:journals/corr/abs-2305-11746}. In contrast, they  are vulnerable to increased amount of pre-trained languages in multilingual setting \citep{DBLP:conf/acl/ConneauKGCWGGOZ20}.
Subsequently, familial LLMs trained on different scales of monolingual data are proved to be viscous \citep{DBLP:journals/corr/abs-2303-16104}, as the source of \textit{oscillatory hallucination} pathology.

\noindent $\bullet$ \textbf{Question and Answer.}
Imperfect responses suffer from flawed external knowledge, knowledge recall cues and reasoning instruction \citep{DBLP:journals/corr/abs-2304-10513}.
For example, LLMs are mostly unable to avoid answering when provided with no relevant information, instead provide incomplete and plausible answers \citep{DBLP:journals/corr/abs-2307-16877}.
In additon to external knowledge, memorized information without accurate, reliable and accessible source also contributes to different types of hallucinations \citep{DBLP:journals/corr/abs-2307-15343}.
Though scaling laws suggest that perplexity on the training distribution is positively correlated with parameter size,
\citep{DBLP:conf/acl/LinHE22} further discovers that scaling up models should increase the rate of imitative falsehoods.

\noindent $\bullet$ \textbf{Dialog System.}
Some studies view dialogue models as unobtrusive imitators, which simulates the distributional properties of data instead of generating faithful output.
For example, \textit{Uncooperativeness responses} \citep{DBLP:conf/naacl/DziriMYZR22} originating from discourse phenomena inclines to output an exact copy of the entire evidence. 
\citet{DBLP:conf/emnlp/DasSS22} reports more nuanced hallucinations in KG-grounded dialogue systems as analyzed through human feedback.
Similarly, \texttt{FaithDial} \citep{DBLP:journals/tacl/DziriKMZYPR22}, \texttt{BEGIN} \citep{DBLP:journals/tacl/DziriRLR22}, \texttt{MixCL} \citep{DBLP:conf/aaai/SunSGRRR23} all implement experiments  on the \texttt{WoW} dataset to conduct a meta-evaluation of the hallucination in knowledge grounded dialogue.

\noindent $\bullet$ \textbf{Summarization System.}
Automatically generated abstracts based on LLMs may be fluent, but they still typically lack faithfulness to the source document. 
Compared to the human evaluation of traditional summarization models \citep{DBLP:conf/acl/MaynezNBM20}, the summarizations generated by LLMs can be categorized into two major types: \textit{intrinsic hallucinations} that distort the information present in the document; \textit{extrinsic hallucinations} that provide additional information that cannot be directly attributed to the document \citep{DBLP:journals/corr/abs-2305-13632}. 
Note that extrinsic hallucination as a metrics of factually consistent continuation of inputs in LLMs is given more attention in summarisation systems \citep{DBLP:conf/acl/TamMZKBR23,DBLP:conf/www/ShenLFRBN23}.
Furthermore,
\citet{DBLP:conf/acl/CaoDC22} subdivides extrinsic hallucinations into  \textit{factual} and \textit{non-factual} hallucinations. The former provides additional world knowledge, which may benefit comprehensive understanding.

\noindent $\bullet$ \textbf{Knowledge Graph with LLMs.}
Despite the promising progress in knowledge-based text geneartion, it encounters \textit{intrinsic hallucinations} inherent to the process where the generated text not only covers the input information but also incorporates redundant details derived from its internal memorized knowledge \citep{DBLP:journals/corr/abs-2307-14712}.
To address this, \citet{DBLP:journals/corr/abs-2306-09296} establish a distinction between correctly generated knowledge and \textit{knowledge  hallucinations}  in terms of knowledge creation.
Notably, the \textit{Virtual Knowledge Extraction}  \citep{DBLP:journals/corr/abs-2305-13168} underscores the potential generalization capabilities of LLMs in the realms of constructing and inferring from knowledge graphs.
\citet{DBLP:journals/corr/abs-2308-02357} further empower LLMs to produce interpretable fact-checks through a neural symbolic approach. 
Based on their fidelity to the source, hallucinations are defined as \textit{subject hallucination}, \textit{relation hallucination}, and \textit{object hallucination}.

\noindent $\bullet$ \textbf{Cross-modal System.}
Augmented by the superior language capabilities of LLMs, performance of cross-modal tasks achieves promising progress \citep{DBLP:journals/corr/abs-2304-10592,DBLP:journals/corr/abs-2304-08485}.
However, despite replacing the original language encoder with LLMs, Large Visual Language Models (LVLMs) \citep{DBLP:conf/icml/WangYMLBLMZZY22} still generate object descriptions that not present in the target image, denoted as \textit{object hallucinations} \citep{DBLP:journals/corr/abs-2305-10355}.
Especially, the various failure cases could be typically found in Visual Question Answering \citep{DBLP:journals/corr/abs-2305-10355}, Image Captioning \citep{DBLP:conf/wacv/BitenGK22,DBLP:journals/corr/abs-2305-07021,DBLP:journals/corr/abs-2305-12943} and Report Generation \citep{DBLP:journals/corr/abs-2307-14634} etc. cross-modal tasks.

\section{Hallucination Detection}
\label{detection}

Conventional hallucination detection mainly depends on task-specific metrics, such as ROUGE and BLEU to evaluate the information overlap between source and target texts in summarization tasks \citep{DBLP:conf/naacl/PagnoniBT21}, while knowledge F1 to estimate the knowledge-aware ability of response generation \citep{DBLP:conf/naacl/LiPSMLYG22}. 
These metrics focus on measuring \textit{faithfulness} of references and fail to provide an assessment of \textit{factualness}.
Despite some reference-free works are proposed, plugin-based methods \citep{DBLP:conf/emnlp/DongWV22} suffer from world knowledge limitation.
QA-based matching metrics \citep{DBLP:conf/acl/DurmusHD20} lack knowledge completeness of source information. 
NLI-based methods \citep{DBLP:journals/tacl/DziriRLR22} are unable to support finer-grained hallucination checking as they are sentence-level, besides entailment and hallucination problems are not equivalent. 
As the paradigm shift in hallucination detection arising from the rapid development of LLMs, we present a novel taxonomy in Fig~\ref{fig:detection} and introduce each category in following sections.

\noindent $\bullet$ \textbf{Inference Classifier.}
The most straightforward strategy involves adopting classifiers to assess the likelihood of hallucinations. 
Concretely, given a question $\mathcal{Q}$ and an answer $\mathcal{A}$, an inferential classifier $\mathcal{C}$ can be asked to determine whether the answer contains hallucinatory content $\mathcal{H}$ via computing $p(\mathcal{H})=\mathcal{F}_\mathcal{C}(\mathcal{Q}, \mathcal{A})$.
Therefore, \citet{DBLP:conf/www/ShenLFRBN23} employs the state-of-the-art LLMs to do end-to-end text generation of detection results.
Some other studies \citep{DBLP:journals/corr/abs-2305-11747} found that adding chains of thought precede output may intervene in the final judgement, whereas retrieving the knowledge properly resulted in gains.
Furthering this concept, the hinted classifer and explainer \citep{DBLP:conf/www/ShenLFRBN23}, used to generate intermediate process labels and high-quality natural language explanations, were demonstrated to enhance the final predicted class from a variety of perspectives.
Subsequently, \citet{DBLP:conf/acl/TamMZKBR23} suggests adopting a different classifier model to the generated model, contributing to easier judgement of factual consistency.
For radiology report generation,  binary classifiers \citep{DBLP:journals/corr/abs-2307-14634} can be leveraged to measure the reliability by combining image and text embedding.

\noindent $\bullet$ \textbf{Uncertainty Metric.}
It is important to examine the correlation between the hallucination metric and the quality of output from a variety of perspectives.
One intuitive approach is to employ the probabilistic output of the model itself, as \texttt{ASTSN} \citep{DBLP:journals/corr/abs-2307-03987} calculates the models' uncertainty about the identified concepts by utilising the logit output values.
Similarly, \texttt{BARTSCORE} \citep{DBLP:conf/nips/YuanNL21} employs a universal notion that models trained to convert generated text to reference output or source text will score higher when the generated text is superior.
It is an unsupervised metric that supports the addition of appropriate prompts to improve the measure design, without human judgement to train.
Furthermore, \texttt{KoK} \citep{DBLP:journals/corr/abs-2305-13712} based on the work of \citet{DBLP:conf/emnlp/PeiJ21} evaluates answer uncertainty from three categories, i.e., subjectivity, hedges and text uncertainty.
However, \texttt{SLAG} \citep{DBLP:conf/eacl/HaseDCLKSBI23} measures consistent factual beliefs in terms of paraphrase, logic, and entailment.
In addition to this, \texttt{KLD}  \citep{DBLP:journals/corr/abs-2306-06264} combines information theory-based metrics (e.g., entropy and KL-divergence) to capture knowledge uncertainty.
Beside expert-stipulated programmatic supervision, \texttt{POLAR} \citep{zhao2023llm} introduces Pareto optimal learning assessed risk score for estimating the confidence level of a response.

\begin{figure}
    \centering
    \resizebox{.49\textwidth}{!}{
    \includegraphics{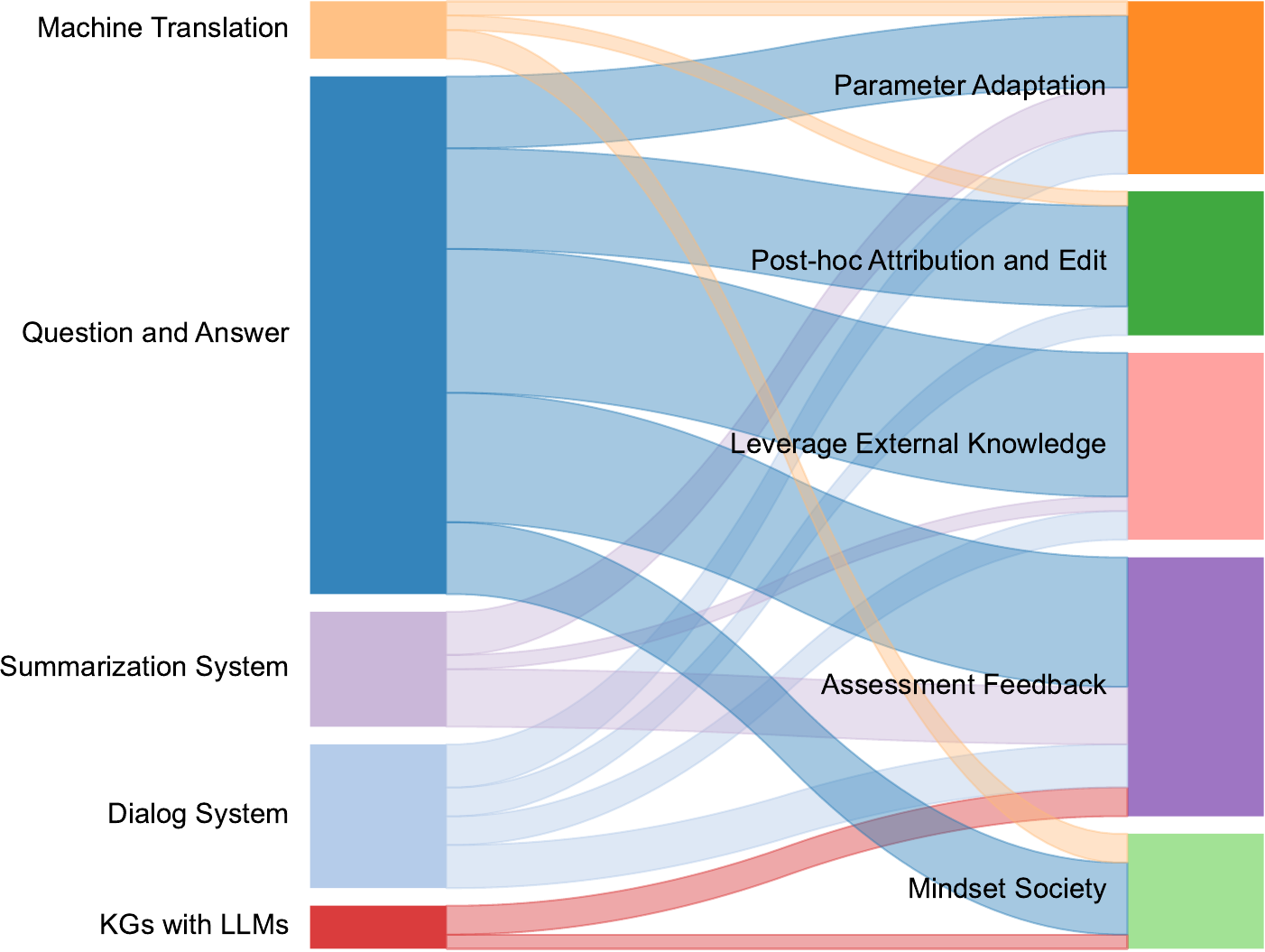}}
    \caption{Sankey diagram of hallucination correction methods with different mainstream NLP tasks.}
    \label{fig:sankey_figure}
\end{figure}

\begin{figure*}[!t]
  \centering
  \resizebox{\textwidth}{!}{
\begin{forest}
  forked edges,
  for tree={
  grow=east,
  reversed=true,
  anchor=base west,
  parent anchor=east,
  child anchor=west,
  base=left,
  font=\small,
  rectangle,
  draw=hiddendraw,
  rounded corners,align=left,
  minimum width=6em,
    edge+={darkgray, line width=1pt},
s sep=1pt,
inner xsep=1pt,
inner ysep=3pt,
ver/.style={rotate=90, child anchor=north, parent anchor=south, anchor=center},
  },
  where level=1{text width=5em,font=\scriptsize,}{},
  where level=2{text width=4em,font=\scriptsize,}{},
  [Hallucination \\ Correction   , ver 
        [Parameter \\ Adaptation  
            [    Factual-Nucleus~\citep{DBLP:conf/nips/LeePXPFSC22}{,}
            CLR~\citep{DBLP:conf/aaai/SunSGRRR23}{,}
            Edit-TA~\citep{DBLP:conf/iclr/IlharcoRWSHF23}{,}
            EWR\\~\citep{DBLP:journals/corr/abs-2303-17574}{,}
            PURR~\citep{DBLP:journals/corr/abs-2305-14908}{,}
            mmT5~\citep{DBLP:journals/corr/abs-2305-14224}{,}
            HISTALIGN\\~\citep{DBLP:journals/corr/abs-2305-04782}{,}
            TYE~\citep{DBLP:journals/corr/abs-2305-14739}{,} 
            ALLM~\citep{DBLP:journals/corr/abs-2305-04757}{,}
            Inference-Time\\~\citep{DBLP:journals/corr/abs-2306-03341}{,}
            TRAC~\citep{DBLP:journals/corr/abs-2307-04642}{,}
            EasyEdit~\citep{DBLP:journals/corr/abs-2308-07269},
            ,leaf,text width=25em]
        ]
        [Post-hoc Attribution \\ and Edit Technology
            [ NP-Hunter~\citep{DBLP:conf/emnlp/DziriMZB21}{,}
            CoT~\citep{DBLP:conf/nips/Wei0SBIXCLZ22}{,}
            ORCA~\citep{DBLP:journals/corr/abs-2205-12600}{,}
            RR\\~\citep{DBLP:journals/corr/abs-2301-00303}{,}
            TRAK~\citep{DBLP:conf/icml/ParkGILM23}{,}
            Data-Portraits~\citep{DBLP:journals/corr/abs-2303-03919}{,}\\
            Self-Refine~\citep{DBLP:journals/corr/abs-2303-17651}{,}
            Reflexion~\citep{DBLP:journals/corr/abs-2303-11366}{,} 
            QUIP~\citep{DBLP:journals/corr/abs-2305-13252}{,}\\
            Verify-and-Edit~\citep{DBLP:conf/acl/ZhaoLJQB23},
            ,leaf,text width=25em]
        ]
        [Leverage External \\ Knowledge
            [ RETRO~\citep{DBLP:conf/icml/BorgeaudMHCRM0L22}{,}
            IRCoT~\citep{DBLP:conf/acl/TrivediBKS23}{,}
            POPQA~\citep{DBLP:conf/acl/MallenAZDKH23}{,} \\
            LLM-AUGMENTER~\citep{DBLP:journals/corr/abs-2302-12813}{,}
            In-Context RALM~\citep{DBLP:conf/acl/TamMZKBR23}{,}
            GeneGPT\\~\citep{DBLP:journals/corr/abs-2304-09667}{,}
            cTBL~\citep{DBLP:conf/candc/DingSMC23}{,}
            CoK~\citep{DBLP:journals/corr/abs-2305-13269}{,} 
            FLARE ~\citep{DBLP:journals/corr/abs-2305-06983}{,}\\
            Gorilla~\citep{DBLP:journals/corr/abs-2305-15334}{,}
            RETA-LLM~\citep{DBLP:journals/corr/abs-2306-05212}{,}
            KnowledGPT~\citep{DBLP:journals/corr/abs-2308-11761},
            ,leaf,text width=25em]
        ]
        [Assessment \\ Feedback
            [ LSHF~\citep{DBLP:conf/nips/StiennonO0ZLVRA20}{,}
            TLM~\citep{DBLP:journals/corr/abs-2203-11147}{,}
            BRIO~\citep{DBLP:conf/acl/LiuLRN22}{,}
            LM-know\\~\citep{DBLP:journals/corr/abs-2207-05221}{,} 
            Chain-of-Hindsight~\citep{DBLP:journals/corr/abs-2302-02676}{,}
            ZEROFEC~\citep{DBLP:conf/acl/HuangCJ23}{,}\\
            CRITIC~\citep{DBLP:journals/corr/abs-2305-11738}{,}
            VIVID~\citep{DBLP:journals/corr/abs-2305-12943}{,}
            LMH-Snowball~\citep{DBLP:journals/corr/abs-2305-13534}{,}\\
            MixAlign~\citep{DBLP:journals/corr/abs-2305-13669}{,} 
            REFEED~\citep{DBLP:journals/corr/abs-2305-14002}{,}
            PaD~\citep{DBLP:journals/corr/abs-2305-13888}{,}
            ALCE \\~\citep{DBLP:journals/corr/abs-2305-14627}{,}
            Do-LLM-Know~\citep{DBLP:journals/corr/abs-2305-18248}{,}
            CRL~\citep{DBLP:conf/acl/DixitWC23},
            ,leaf,text width=25em]
        ]
        [Mindset Society
            [ HLMTM~\citep{DBLP:journals/corr/abs-2303-16104}{,}
            Multiagent-Debate \citep{DBLP:journals/corr/abs-2305-14325}{,}
            MAD~\citep{DBLP:journals/corr/abs-2305-19118}{,} \\
            FORD~\citep{DBLP:journals/corr/abs-2305-11595}{,}
            LM-vs-LM~\citep{DBLP:journals/corr/abs-2305-13281}{,}
            PRD~\citep{DBLP:journals/corr/abs-2307-02762}{,}
            SPP\\~\citep{DBLP:journals/corr/abs-2307-05300},
            ,leaf,text width=25em]
        ]
    ]
\end{forest}
}
\caption{Taxonomy of Hallucination Correction.}
\label{Taxonomy of Hallucination Correction}
\end{figure*}
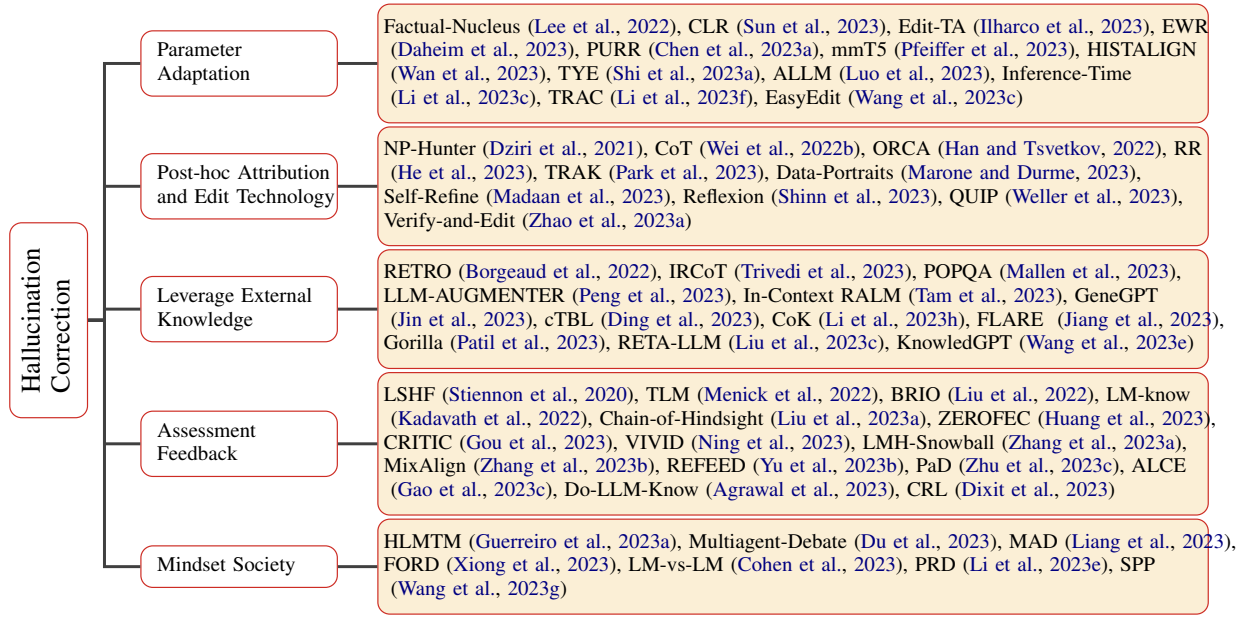

\noindent $\bullet$ \textbf{Self-Evaluation.}
To self-evaluate is challenging since the model might be overconfident about its generated samples being correct.
The motivating idea of \texttt{SelfCheckGPT} \citep{DBLP:journals/corr/abs-2303-08896} is to use the ability of the LLMs themselves to sample multiple responses and identify fictitious statements by measuring the consistency of information among responses.
\citet{DBLP:journals/corr/abs-2207-05221} further illustrates that both the increase in size and the demonstration of assessment can improve self-assessment.
Beyond repetitive multiple direct queries, \citet{DBLP:journals/corr/abs-2305-18248} uses open-ended indirect queries and compares their answers to each other for an agreed-upon score outcome.
\texttt{Self-Contradictory} \citep{DBLP:journals/corr/abs-2305-15852} imposes appropriate constraints on the same LLM to generate pairs of sentences triggering self-contradictions, which prompt the detection.
In contrast, Polling-based querying \citep{DBLP:journals/corr/abs-2305-10355} reduce the complexity of judgement by randomly sampling query objects.
Besides, \texttt{Self-Checker} \citep{DBLP:journals/corr/abs-2305-14623} decomposes complex statements into multiple simple statements, fact-checking them one by one.
However, \citet{DBLP:journals/corr/abs-2305-13281} introduces two LLMs interacting cross examination to drive the complex fact-checking reasoning process.

\noindent $\bullet$ \textbf{Evidence Retrieval.}
Evidence retrieval accomplishes factual detection by retrieving supporting evidence related to hallucinations.
To this end, Designing a claim-centric pipeline allows for a question-retrieve-summary chain to effectively collect original evidence \citep{DBLP:journals/corr/abs-2305-11859, DBLP:journals/corr/abs-2306-13781}.
Consequently, \texttt{FActScore} \citep{DBLP:journals/corr/abs-2305-14251} calculates the percentage of atomic facts supported by the given knowledge source.
Towards adapting the tasks that users in interaction with generative models, \texttt{FacTool} \citep{DBLP:journals/corr/abs-2307-13528} proposes to integrate a variety of tools into a task-agnostic and domain-agnostic detection framework, in order to assemble evidence about the authenticity of the generated content.

\section{Hallucination Correction}
\label{correction}

In this section, we delve into the methods to correct hallucination in terms of different aspects,  i.e. \textit{Parameter Adaptation}, \textit{Post-hoc Attribution and Edit Technology}, \textit{Leverage External Knowledge}, \textit{Assessment Feedback}, \textit{Mindset Society}. As shown in Figure \ref{fig:sankey_figure}, these hallucination correction paradigms have demonstrated strong dominance in many mainstream NLP tasks. 
Note that these methods are not entirely orthogonal but could complement each other as required by the tasks in practical applications.
In the following sections,  we will introduce each methods as shown in Figure \ref{Taxonomy of Hallucination Correction}.

\noindent $\bullet$ \textbf{Parameter Adaptation.}
Parameters in LLMs store biases learned in pre-training, are often unaligned with user intent.
A cutting-edge strategy is to guide effective knowledge through parameter conditioning, editing, and optimisation.
For example, \texttt{CLR}~\citep{DBLP:conf/aaai/SunSGRRR23} optimises to reduce the generation probability of negative samples at span level utilising contrastive learning parameters.
While introducing contextual knowledge background that contradicts the model's intrinsic prior knowledge, \texttt{TYE}~\citep{DBLP:journals/corr/abs-2305-14739} effectively reduces the weight of prior knowledge through context-aware decoding method.
Besides, \texttt{PURR}~\citep{DBLP:journals/corr/abs-2305-14908} corrupts noise into the text, fine-tune compact editors, and denoise by merging relevant evidence.
To introduce additional cache component, \texttt{HISTALIGN}~\citep{DBLP:journals/corr/abs-2305-04782} discovers that its hidden state is not aligned with the current hidden state, and proposes sequence information contrastive learning to improve the reliability of memory parameters.
Consequently, \texttt{Edit-TA}~\citep{DBLP:conf/iclr/IlharcoRWSHF23} mitigates the biases learnt in pre-training from a task algorithm perspective.
An intuition behind it is that parameter variations learnt through negative example tasks could be perceived through weight variances.
However as this fails to take the importance of different negative examples into account, therefore \texttt{EWR}~\citep{DBLP:journals/corr/abs-2303-17574} proposes Fisher information matrices to measure the uncertainty of their estimation, which is applied for the dialogue systems to execute a parameter interpolation and remove hallucination.
\texttt{EasyEdit}~\citep{DBLP:journals/corr/abs-2308-07269} summarises methods for parameter editing, while minimising the influence to irrelevant parameter.

An efficient alternative is to identify task-specific parameters and exploit them.
For example, \texttt{ALLM}~\citep{DBLP:journals/corr/abs-2305-04757} aligns the parameter module with task-specific knowledge, and then generates the relevant knowledge as additional context in background augmented prompts.
Similarly, \texttt{mmT5}~\citep{DBLP:journals/corr/abs-2305-14224} utilises language-specific modules during pre-training to separate language-specific information from language-independent information, demonstrating that adding language-specific modules can alleviate the curse of multilinguality.
Instead, \texttt{TRAC}~\citep{DBLP:journals/corr/abs-2307-04642} combines conformal prediction and global testing to augment retrieval-based QA.
The conservative strategy formulation ensures that a semantically equivalent answer to the truthful answer is included in the prediction set.

Another parameter adaptation idea focuses on flexible sampling consistent with user requirements.
For instance, \citet{DBLP:conf/nips/LeePXPFSC22} observes that the randomness of sampling is more detrimental to factuality when generating the latter part of a sentence.
The factual-nucleus sampling algorithm is introduced to keep the faithfulness of the generation while ensuring the quality and diversity.
Besides, \texttt{Inference-Time}~\citep{DBLP:journals/corr/abs-2306-03341} firstly identifies a set of attentional heads with high linear probing accuracy, and then shifts activation in the inference process along the direction associated with factual knowledge.

\noindent $\bullet$ \textbf{Post-hoc Attribution and Edit Technology.}
A source of hallucination is that LLMs may leverage the patterns observed in the pre-training data for inference in a novel form.
Recently, \texttt{ORCA}~\citep{DBLP:journals/corr/abs-2205-12600} reveals problematic patterns in the behaviour of models by probing supporting data evidences from pre-training data.
Similarly, \texttt{TRAK} \citep{DBLP:conf/icml/ParkGILM23} and \texttt{Data-Portraits} \citep{DBLP:journals/corr/abs-2303-03919} analyse whether models plagiarise or reference existing resources by means of data attribution.
\texttt{QUIP} \citep{DBLP:journals/corr/abs-2305-13252} further demonstrates that providing text that has been observed in the pre-training phase can improve the ability of LLMs to generate more factual information.
Furthermore, motivated by the gap between LLMs and human modes of thinking, one intuition is to align the two modes of reasoning.
Thus \texttt{CoT} \citep{DBLP:conf/nips/Wei0SBIXCLZ22} elicits faithful reasoning via a kind of Chain-of-Thought (CoT)  \citep{DBLP:conf/nips/KojimaGRMI22}  prompts.
Similarly, \texttt{RR} \citep{DBLP:journals/corr/abs-2301-00303} retrieves relevant external knowledge based on decomposed reasoning steps obtained from a CoT prompt.
Since LLMs not often produce the best output on the first attempt, \texttt{Self-Refine} \citep{DBLP:journals/corr/abs-2303-17651} implements self-refinement algorithms through iterative feedback and improvement.
\texttt{Reflexion} \citep{DBLP:journals/corr/abs-2303-11366} also employs verbal reinforcement to generate reflective feedback by learning about prior failings.
\texttt{Verify-and-Edit} \citep{DBLP:conf/acl/ZhaoLJQB23} proposes a CoT-prompted verify-and-edit framework, which improves the fidelity of predictions by post-editing the inference chain based on externally retrieved knowledge.
Another source of hallucinations is to describe factual content with incorrect retrievals.
To recify this, \texttt{NP-Hunter} \citep{DBLP:conf/emnlp/DziriMZB21} follows a generate-then-refine strategy whereby a generated response is amended using the KG so that the dialogue systemable to correct potential hallucinations by querying the KG.

\noindent $\bullet$ \textbf{Leverage External Knowledge.}
As an attempt to extend the language model for halucination mitigation, a suggestion is to retrieve relevant documents from large textual databases.
\texttt{RETRO}~\citep{DBLP:conf/icml/BorgeaudMHCRM0L22} splits the input sequence into chunks and retrieves similar documents, while \texttt{In-Context RALM} \citep{DBLP:conf/acl/TamMZKBR23} places the selected document before the input text to improve the prediction.
Furthermore, \texttt{IRCoT} \citep{DBLP:conf/acl/TrivediBKS23} interweaves CoT generation and document retrieval steps to guide LLMs.
Since scaling mainly improves the memory for common knowledge but does not significantly improve the memory for factual knowledge in the long tail,  \texttt{POPQA} \citep{DBLP:conf/acl/MallenAZDKH23} retrieves only non-parametric memories when necessary to improve performance.
\texttt{LLM-AUGMENTER} \citep{DBLP:journals/corr/abs-2302-12813} also bases the responses of LLMs on integrated external knowledge and automated feedback to improve the truthfulness score of the answers.
Another work, \texttt{CoK} \citep{DBLP:journals/corr/abs-2305-13269} iteratively analyses future content of upcoming sentences, and then applies them as a query to retrieve relevant documents for the purposes of re-generating sentences when they contain low confidence tokens.
Similarly, \texttt{RETA-LLM} \citep{DBLP:journals/corr/abs-2306-05212} creates a complete pipeline to assist users in building their own domain-based LLM retrieval systems.
Note that in addition to document retrieval, diverse external knowledge queries coule be assembled into retrieval-augmented LLM systems.
For example, \texttt{FLARE} \citep{DBLP:journals/corr/abs-2305-06983} leverages structured knowledge bases to support complex queries and provide more straightforward factual statements.
Further, \texttt{KnowledGPT} \citep{DBLP:journals/corr/abs-2308-11761} adopts program of thoughts (PoT) prompting, which generates codes to interact with knowledge bases. 
While \texttt{cTBL} \citep{DBLP:conf/candc/DingSMC23} proposes to enhance LLMs with tabular data in conversation settings.
Besides, \texttt{GeneGPT} \citep{DBLP:journals/corr/abs-2304-09667} demonstrates that expertise can be accessed more easily and accurately by detecting and executing API calls through contextual learning and augmented decoding algorithms.
To support potentially millions of ever-changing APIs, \texttt{Gorilla} \citep{DBLP:journals/corr/abs-2305-15334} explores self-instruct fine-tuning and retrieval for efficient API exploitation.

\noindent $\bullet$ \textbf{Assessment Feedback.}
As language models become more sophisticated, evaluation feedback can significantly improve the quality of generated text, as well as reduce the appearance of hallucinations.
To realise this concept,
\texttt{LSHF} \citep{DBLP:conf/nips/StiennonO0ZLVRA20} predicts human-preferred summarizations through model and employs this as the reward function to fine-tune the summarization strategy using reinforcement learning.
However, this approach builds on models crafted by human annotators, which makes them inefficient in terms of data utilization.
Therefore, \texttt{TLM} \citep{DBLP:journals/corr/abs-2203-11147} proposes to improve the reliability of the system by selecting a few questions to refuse to answer, which significantly improves the reliability of the system, through reinforcement learning from human preferences.
Whereas reinforcement learning often suffers from imperfect reward functions and relies on challenging optimisation.
Thus, \texttt{Chain-of-Hindsight} \citep{DBLP:journals/corr/abs-2302-02676} converts feedback preferences into sentences, which are then fed into models with fine-tuning for enhanced language comprehension.

In addition to enabling the model to learn directly from the feedback of factual metrics in a sample-efficient manner \citep{DBLP:conf/acl/DixitWC23}, it is also important to build in a self-evaluation function of the model to filter candidate generated texts.
For example, \texttt{BRIO} \citep{DBLP:conf/acl/LiuLRN22} empowers summarization model assessment, estimating probability distributions of candidate outputs to rate the quality of candidate summaries.
While \texttt{LM-know} \citep{DBLP:journals/corr/abs-2207-05221} is devoted to investigating whether LLMs can evaluate the validity of their own claims by detecting the probability that they know the answer to a question.
Consequently, \texttt{Do-LLM-Know} \citep{DBLP:journals/corr/abs-2305-18248} queries exclusively with black-box LLMs, and the results of queries repeatedly generated multiple times are compared with each other to pass consistency checks.
Besides, the black-box LLM is augmented with a plug-and-play retrieval module \citep{ DBLP:journals/corr/abs-2302-02676, DBLP:conf/acl/HuangCJ23} to generate feedback could improve the model response.

As missing citation quality evaluation affects the final performance, \texttt{ALCE} \citep{DBLP:journals/corr/abs-2305-14627} employs a natural language reasoning model to measure citation quality and extends the integrated retrieval system.
Similarly, \texttt{CRITIC} \citep{DBLP:journals/corr/abs-2305-11738} proposes to interact with appropriate tools to assess certain aspects of the text, and then to modify the output based on the feedback obtained during the verification process.
Note that automated error checking can also utilise LLMs to generate text that conforms to tool interfaces.
\texttt{PaD} \citep{DBLP:journals/corr/abs-2305-13888} distills the LLMs with a synthetic inference procedure, and the synthesis program obtained can be automatically compiled and executed by an explainer.
Further, iterative refinement processes are validated to effectively identify appropriate details \citep{DBLP:journals/corr/abs-2305-12943,DBLP:journals/corr/abs-2305-13669,DBLP:journals/corr/abs-2305-14002}, and can stop early invalid reasoning chains, beneficially reducing the phenomenon of hallucination snowballing \citep{DBLP:journals/corr/abs-2305-13534}.

\noindent $\bullet$ \textbf{Mindset Society.}
Human intelligence thrives on the concept of cognitive synergy, where collaboration between different cognitive processes produces better results than isolated individual cognitive processes. 
"Society of minds" \citep{minsky1988society} is believed have the potential to significantly improve the performance of LLMs and pave the way for consistency in language production and comprehension.
For the purpose of addressing hallucinations in large-scale multilingual models across different translation scenarios, \texttt{HLMTM} \citep{DBLP:journals/corr/abs-2303-16104} proposes a hybrid setting in which other translation systems can be requested to act as a back-up system when the original system is hallucinating.
Consequently, \texttt{Multiagent-Debate} \citep{DBLP:journals/corr/abs-2305-14325} employs multiple LLMs in several rounds to propose and debate their individual responses and reasoning processes to reach a consensus final answer.
As a result of this process, the models are encouraged to construct answers that are consistent with both internal criticisations and responses from other agents. 
Before a final answer is presented, the resultant community of models can hold and maintain multiple reasoning chains and possible answers simultaneously.
Based on this idea, \texttt{MAD} \citep{DBLP:journals/corr/abs-2305-19118} adds a judge-managed debate process, demonstrating that adaptive interruptions of debate and controlled "tit-for-tat" states help to complete factual debates.
Furthermore, \texttt{FORD} \citep{DBLP:journals/corr/abs-2305-11595} proposes roundtable debates that include more than two LLMs and emphasises that competent judges are essential to dominate the debate.
\texttt{LM-vs-LM} \citep{DBLP:journals/corr/abs-2305-13281} also proposes multi-round interactions between LM and another LM to check the factualness of original statements.
Besides, \texttt{PRD} \citep{DBLP:journals/corr/abs-2307-02762} proposes a peer rank and discussionbased evaluation framework to arrive at a well-recognised assessment result that all peers are in agreement with.
In an effort to maintain strong reasoning, 
\texttt{SPP} \citep{DBLP:journals/corr/abs-2307-05300} utilises LLMs to assign several fine-grained roles, which effectively stimulates knowledge acquisition and reduces hallucinations.

\section{Future Directions}
\label{future work}

Though numerous technical solutions have been proposed in the survey for hallucinations in LLMs, there exist some potential directions:

\noindent $\bullet$ \textbf{Data Construction Management.}
As previously discussed, the style, and knowledge of LLMs is basically learned during model pre-training.
High quality data present promising opportunities for  facilitating the reduction of hallucinations in LLMs \citep{DBLP:conf/emnlp/KirstainL0L22}. 
Inspired by the basic rule of machine learning models: "Garbage  input, garbage  output", \citet{DBLP:journals/corr/abs-2305-11206} proposes the superficial alignment hypothesis, which views alignment as learning to interact with the user. 
The results of simple fine-tuning on a few high-quality samples demonstrate that data quality and diversity outweigh the importance of fine-tuning large-scale instructions \citep{DBLP:journals/corr/abs-2104-08773,DBLP:conf/iclr/WeiBZGYLDDL22,DBLP:conf/iclr/SanhWRBSACSRDBX22} and RLHF \citep{DBLP:journals/corr/abs-2204-05862,DBLP:conf/nips/Ouyang0JAWMZASR22}.
To perform efficiently in knowledge-intensive verticals, we argue that construction of entity-centred fine-tuned instructions \citep{bao2023discmedllm,DBLP:journals/corr/abs-2305-11527,zhu2023ChatMed} is a promising direction that it can combine the structured knowledge and semantic relevance of knowledge graphs to enhance the factuality of generated entity information.
Another feasible proposal is to incorporate a self-curation phase \citep{DBLP:journals/corr/abs-2308-06259} in the instruction construction process to rate the quality of candidate pairs.
During the iteration process, quality evaluation \citep{DBLP:journals/corr/abs-2307-08701} based on manual or automated rule constraints could provide self-correction capacity.

\noindent $\bullet$ \textbf{Downstream Task Alignment.}
Generic LLMs have a certain degree of natural language problem comprehension in a variety of open environments.
However, the main problem still remains in the deviation from the application requirements, which leads to emergence of diverse hallucinations.
Thus, downstream task alignment especially built on vertical domain cognition necessitates expanded symbolic reasoning, decomposition and planning of complex tasks, and faithful external knowledge injection.
Specifically, while expert in language processing, LLMs struggle to make breakthroughs in mathematical abilities, a deficiency attributable to the 
textual training objective.
Though some researches for symbolic math word problems \citep{DBLP:conf/acl/GaurS23,DBLP:conf/acl/ZhuWZZ0GZY23} have been proposed, 
enhancing symbolic reasoning and answering numerical questions remains to be explored extensively.
Additionally, for story generation tasks that demand diverse outputs \citep{DBLP:conf/emnlp/YangTPK22,DBLP:conf/acl/YangKPT23}, fascinating storylines are required in addition to avoiding factual contradictions.
Therefore, achieving a balance between faithfulness and creativity in the model inference process remains a crucial  challenge.
Moreover, integrating new knowledge to deal with knowledge-intensive tasks involves handling joint reasoning between the implicit knowledge of LLMs and the explicit knowledge of external knowledge graphs.
There arises a challenge to design knowledge-aware methods to incorporate structured information from the knowledge graph into the pre-training process of LLMs.
Alternatively, the reasoning process is expected to be dynamically injected with knowledge graph information \citep{DBLP:journals/corr/abs-2308-09729}.

The utilization of LLMs as an evaluation tool is a burgeoning application, but limited by the size of models, the effect of instruction adjustments, and the different forms of inputs \citep{DBLP:journals/corr/abs-2305-18248}.
Note attempts of LLMs to act as judges for scoring have to overcome all kinds of biases induced by position, verbosity, self-enhancement \citep{DBLP:journals/corr/abs-2306-05685,berglund2023taken,wang2023large}.
Therefore, we forecast that future research about designing task-specific mechanisms for analysing and correcting processes of emerging downstream tasks is an area deserving long-term attention.

\noindent $\bullet$ \textbf{Reasoning Mechanism Exploitation.}
The emerging CoT technique \citep{DBLP:conf/nips/Wei0SBIXCLZ22} stimulates the emergent reasoning ability of LLMs by imitating intrinsic stream of thought.
Constructing a logically intermediate reasoning step has been proved to significantly improve the problem-solving ability.
Recently, A primary improvement is Self-consistency with CoT (CoT-SC) \citep{DBLP:journals/corr/abs-2305-10601}, which is a method for generating multiple CoT options and then selecting the optimal result as feedback.
Further, Tree of Thoughts (ToT) \citep{DBLP:journals/corr/abs-2305-10601} introduces a strict tree architecture into the thought process, which facilitates the development with different paths of thought and provides a novel roll-back function.
Since previous methods have no storages for intermediate results, Cumulative Reasoning (CR) \citep{DBLP:journals/corr/abs-2308-04371} uses LLMs in a cumulative and iterative manner to simulate human thought processes, and decompose the task into smaller components.
However, the actual thinking process creates a complex network of ideas, as an example, people could explore a particular chain of reasoning, backtrack or start a new chain of reasoning.
In particular when aware that an idea from a previous chain of reasoning can be combined with the currently explored idea, they could be merged into a new solution.
More excitingly, Graph of Thoughts(GoT) \citep{DBLP:journals/corr/abs-2308-04371} extends the dependencies between thoughts by constructing vertices with multiple incoming edges to aggregate arbitrary thoughts.
In addition, Program-aided language models (PAL) \citep{DBLP:conf/icml/GaoMZ00YCN23} and Program of Thoughts prompting (PoT) \citep{DBLP:journals/corr/abs-2211-12588} introduce programming logic into the language space \citep{bi2023programofthoughts}, expanding the ability to invoke external explainers.
As a summary, we believe that research based on human cognition helps to provide brilliant and insightful insights into the analysis of hallucinations, such as Dual Process Theory \citep{frankish2010dual}, Three layer mental model \citep{stanovich2011rationality}, Computational Theory of Mind \citep{piccinini2004first}, and Connectionism \citep{thorndike1898animal}.

\noindent $\bullet$ \textbf{Multi-modal Hallucination Survey.}
It has become a community consensus to establish powerful Multimodal Large Language Models (MLLMs) \citep{DBLP:conf/icml/0008LSH23,DBLP:journals/corr/abs-2305-06500,DBLP:journals/corr/abs-2304-14178}  by taking advantage of excellent comprehension and reasoning capabilities of LLMs.
\citet{DBLP:journals/corr/abs-2305-10355} confirms the severity of hallucinations in MLLM by object detecting and polling-based querying.
The results indicate that the models are highly susceptible to object hallucination, and the generated description does not match the target image.
Besides, \citet{DBLP:journals/corr/abs-2308-03729} that MLLMs have limited multimodal reasoning ability as well as dependence on spurious cues. 
Though current study \citep{DBLP:journals/corr/abs-2306-13549} provides a broad overview of MLLMs, the causation of hallucinations has not been comprehensively investigated.
The hallucinations in LLMs come mainly from misknowledge in the training data, whereas the challenge of MLLMs lies in accurately relaying the abstract visual encoding into the semantic space.
Existing MLLMs are fine-tuned with instructions to make their target outputs follow human intentions.
However, misalignment between visual and textual modes may lead to biased distribution.
Further, the lack of visual constraints results in a serious problem of hallucination in MLLMs.
Thus a potential improvement is to penalise deviating attention to images \citep{DBLP:journals/corr/abs-2308-15126} or to enhance understanding of visual common sense.
In terms of fine-grained visual and textual modal alignment, focusing on local features of images and corresponding textual descriptions can provide faithful modal interactions.
In addition, the performance of some MLLMs such as MiniGPT-4 \citep{DBLP:journals/corr/abs-2304-10592} is highly dependent on the choice of prompts and requires careful selection.
Note that a controlled trade-off between diversity and hallucinations is needed for user convenience.
In the future, as more sophisticated multi-model applications emerge, improving MLLMs reasoning paths is also a promising research direction.

\section{Conclusion and Vision}
\label{conclusion}

In this paper, we provide an overview of hallucinations in LLMs with new taxonomy, theoretical insight, detection methods, correction methods and several future research directions.
Note that it is crucial to ensure we can continuously utilize LLMs in a responsible and beneficial manner, so we explore the causation of hallucinations and taxonomy in task axes to analyse potential directions for improvement.
In the future, we envision a more potent synergy between LLMs and external knowledge bases, resulting in a credible interactive system with the dual-wheel drive.
We hope that sophisticated and efficient detection methods are proposed to contribute further to improving the performance of LLMs.
Furthermore, we hope that community maintains a proactive attitude towards mitigating the effects of hallucinations.
With creative corrective methods proposed for various aspects, LLMs will provide human with reliable and secure information in broad application scenarios.

\bibliography{reference}
\bibliographystyle{acl_natbib}
\newpage

\end{document}